\documentclass[11pt,a4paper]{article}
\usepackage[hyperref]{naaclhlt2019}
\usepackage{times}
\usepackage{amsmath}
\usepackage{graphicx}
\usepackage{caption}
\usepackage{subcaption}
\usepackage{float}
\usepackage{booktabs}
\usepackage{natbib}
\usepackage{latexsym}
\usepackage{todonotes}
\usepackage{algorithm}
\usepackage{algorithmicx}
\usepackage[noend]{algpseudocode}

\usepackage{url}

\aclfinalcopy 

\title{Understanding Dataset Design Choices for Multi-hop Reasoning}

\author{Jifan Chen \and Greg Durrett \\
  The University of Texas at Austin \\
  {\tt \{jfchen,gdurrett\}@cs.utexas.edu}}

\date{}

\begin{document}
\maketitle
\begin{abstract}
  Learning multi-hop reasoning has been a key challenge for reading comprehension models, leading to the design of datasets that explicitly focus on it. Ideally, a model should not be able to perform well on a multi-hop question answering task without doing multi-hop reasoning. In this paper, we investigate two recently proposed datasets, WikiHop~\cite{welbl2018constructing} and HotpotQA~\cite{yang2018hotpotqa}. First, we explore sentence-factored models for these tasks; by design, these models cannot do multi-hop reasoning, but they are still able to solve a large number of examples in both datasets. Furthermore, we find spurious correlations in the unmasked version of WikiHop, which make it easy to achieve high performance considering only the questions and answers. Finally, we investigate one key difference between these datasets, namely span-based vs.~multiple-choice formulations of the QA task. Multiple-choice versions of both datasets can be easily gamed, and two models we examine only marginally exceed a baseline in this setting. Overall, while these datasets are useful testbeds, high-performing models may not be learning as much multi-hop reasoning as previously thought.
\end{abstract}

\section{Introduction}
Question answering from text \cite{richardson2013mctest,hill2015goldilocks,hermann2015teaching,rajpurkar2016squad} is a key challenge problem for NLP that tests whether models can extract information based on a query. However, even sophisticated models that perform well on QA benchmarks \cite{seo2016bidirectional,shen2017reasonet,yu2018qanet} may only be doing shallow pattern matching of the question against the supporting passage \cite{weissenborn2017making}. More recent work~\cite{kumar2016ask,joshi2017triviaqa,welbl2018constructing} has emphasized gathering information from different parts of a passage to answer the question, leading to a number of models designed to do \emph{multi-hop reasoning}. Two recent large-scale datasets have been specifically designed to test multi-hop reasoning: WikiHop~\cite{welbl2018constructing} and HotpotQA~\cite{yang2018hotpotqa}.

In this paper, we seek to answer two main questions. First, although the two datasets are explicitly constructed for multi-hop reasoning, do models really need to do multi-hop reasoning to do well on them? Recent work has shown that large-scale QA datasets often do not exhibit their advertised properties \cite{chen2016thorough,kaushik2018much}. We devise a test setting to see whether multi-hop reasoning is necessary: can a model which treats each sentence independently select the sentence containing the answer? This provides a rough estimate of the fraction of questions solvable by a non-multi-hop system. Our results show that more than half of the questions in WikiHop and HotpotQA do not require multi-hop reasoning to solve. Surprisingly, we find that a simple baseline which ignores the passage and only uses the question and answer can achieve strong results on WikiHop and a modified version of HotpotQA, further confirming this view.

Second, we study the nature of the supervision on the two datasets. One critical difference is that HotpotQA is span-based (the answer is a span of the passage) while WikiHop is multiple-choice. How does this difference affect learning and evaluation of multi-hop reasoning systems? We show that a multiple-choice version of HotpotQA is vulnerable to the same baseline that performs well on WikiHop, showing that this distinction may be important from an evaluation standpoint. Furthermore, we show that a state-of-the-art model, BiDAF++, trained on span-based HotpotQA and adapted to the multiple-choice setting outperforms the same model trained natively on the multiple-choice setting. However, even in the span-based setting, the high performance of the sentence-factored models raises questions about whether multi-hop reasoning is being learned.

Our conclusions are as follows: (1) Many examples in both WikiHop and HotpotQA do not require multi-hop reasoning to solve, as the sentence-factored model can find the answers. (2) On WikiHop and a multiple-choice version of HotpotQA, a no context baseline does very well. (3) Span-based supervision provides a harder testbed than multiple choice by having more answers to choose from, but given the strong performance of the sentence-factored models, it is unclear whether any of the proposed models are doing a good job at multi-hop reasoning in any setting.


\section{Datasets}

\paragraph{WikiHop} \citet{welbl2018constructing} introduced this English dataset specially designed for text understanding across multiple documents. The dataset consists of 40k+ questions, answers, and passages, where each passage consists of several documents collected from Wikipedia. Questions are posed as a query of a relation $r$ followed by a head entity $h$, with the task being to find the tail entity $t$ from a set of entity candidates $E$. Annotators followed links between documents and were required to use multiple documents to get the answer.

\paragraph{HotpotQA} \citet{yang2018hotpotqa} proposed a new dataset with 113k English Wikipedia-based question-answer pairs. The questions are diverse, falling into several categories, but all require finding and reasoning over multiple supporting documents to answer. Models should choose answers by selecting variable-length spans from these documents. Sentences relevant to finding the answer are annotated in the dataset as ``supporting facts'' so models can use these at training time as well.

\section{Probing Multi-hop Datasets}

In this section, we seek to answer whether multi-hop reasoning is really needed to solve these two multi-hop datasets.

\begin{table}[t]
\small
\centering
\begin{tabular}{l|c |c | c}
\toprule
Method & Random & Factored & Factored BiDAF \\ \midrule
WikiHop & 6.5 & 60.9 & 66.1 \\
HotpotQA & 5.4 & 45.4 & 57.2 \\
SQuAD & 22.1 & 70.0 & 88.0 \\
\bottomrule
\end{tabular}
\caption{The accuracy of our proposed sentence-factored models on identifying answer location in the development sets of WikiHop, HotpotQA and SQuAD. \emph{Random}: we randomly pick a sentence in the passage to see whether it contains the answer. \emph{Factored} and \emph{Factored BiDAF} refer to the models of Section~\ref{section:decoupled}. As expected, these models perform better on SQuAD than the other two datasets, but the model can nevertheless find many answers in WikiHop especially.}
\vspace{-0.5cm}
\label{tab:decouple-test}
\end{table}

\subsection{Sentence-Factored Model Test}
\label{section:decoupled}

\begin{figure}[t]
    \centering
    \includegraphics[width=0.5\textwidth]{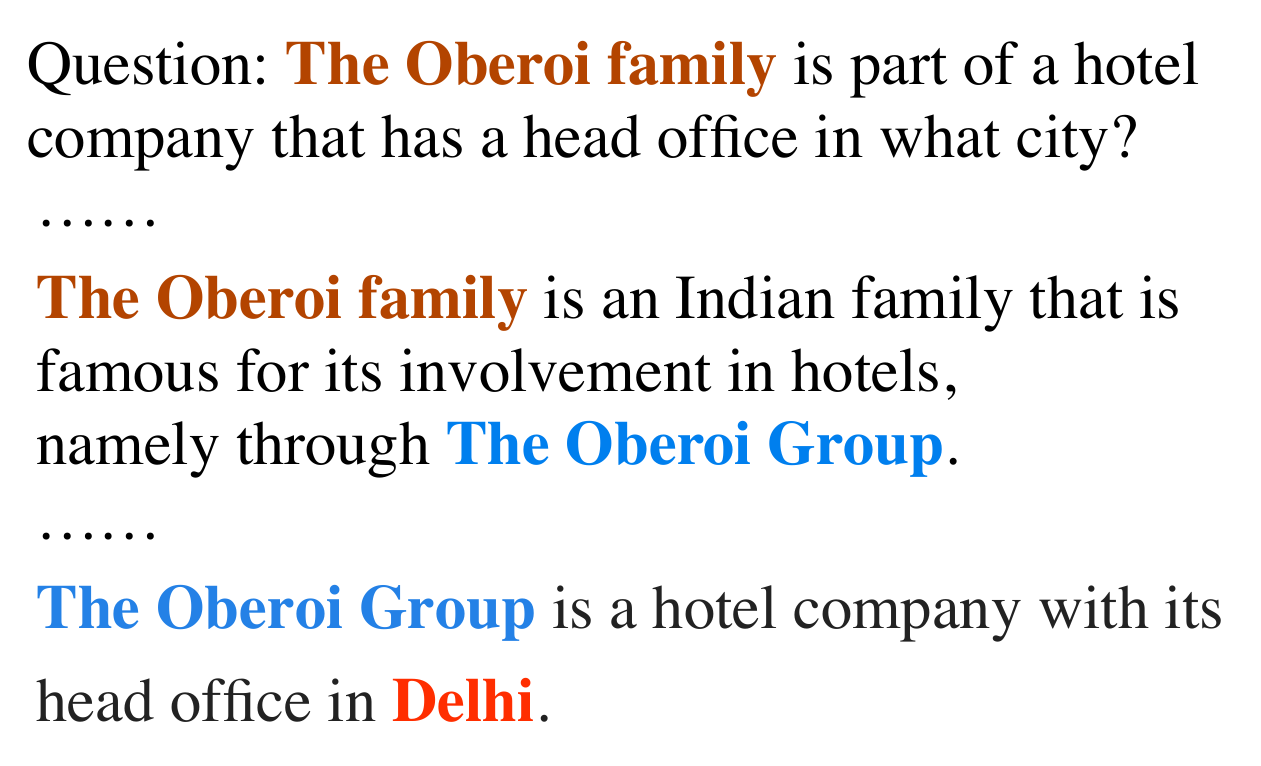}
    \caption{An example from the HotpotQA dev set. Here, a model should have to form a reasoning chain \emph{Oberoi family} $\rightarrow$ \emph{Oberoi Group} $\rightarrow$ \emph{Delhi} to arrive at the answer. However, the sentence containing \emph{Delhi} has a substantial lexical overlap with the question, so strong QA systems can answer it directly.}
    \label{fig:hotpot_example}
    \vspace{-0.5cm}
\end{figure}

If a question requires a multi-hop model, then we should not be able to figure out the answer by only looking at the question and each sentence separately. Based on this idea, we propose a sentence-factored modeling setting, where a model must predict which sentence contains the answer but must score each sentence independently, i.e., without using information from other sentences in this process. Identifying the presence of the answer is generally easier than predicting the answer directly, particularly if a sentence is complicated, and is still sufficient to provide a bound on how strongly multi-hop reasoning is required. Figure~\ref{fig:hotpot_example} shows a typical example from these datasets, where identifying the answer (\emph{Delhi}) requires bridging to an entity not mentioned in the question.

\paragraph{Simple Factored Model} We encode each passage sentence $s_i$ and the question $q$ into a contextual representation $h_{s_i}$ and $h_q$ using a bi-directional GRU~\cite{chung2014empirical}. Then, $S_i = h_{s_i}^\top W h_q$; that is, compute a bilinear product of these representations with trainable weights $W$ to get the score of the $i$th sentence. Finally, let $p_i = \textrm{softmax}_i(S_i)$; softmax over the sentences to get a probability distribution. We maximize the marginal log probability of picking a sentence containing the correct answer: $\log(\sum_{i: s_i \in \mathbf{s^*}} p_i)$, where $\mathbf{s^*}$ is the set of sentences containing the answer. During evaluation, we pick the sentence $s$ with the highest score and treat it as correct if it contains the answer.

\paragraph{Factored BiDAF} We encode the question and each sentence \emph{separately} using bi-GRUs. Then, we generate the question-aware token representation for each token of sentence by using a co-attention layer~\cite{seo2016bidirectional}. Finally, we max-pool over each sentence to get the sentence representation and feed those to a FFNN to compute the sentence score. Training and inference are the same as for the simple model.

We run this test on both datasets as well as SQuAD \cite{rajpurkar2016squad}, where multi-hop reasoning is only needed in a few questions. Results in Table~\ref{tab:decouple-test} indicate that although intentionally created for multi-hop reasoning, for more than half of questions in WikiHop and HotpotQA, we can figure out where the answer is without doing  multi-hop reasoning. This result is initially surprising, but one reason it may be possible is suggested by the example from HotpotQA shown in Figure~\ref{fig:hotpot_example}. We can see that the model could easily figure out the answer sentence without looking at the bridging entities using lexical cues alone. This observation is also in accordance with the work of~\newcite{jansen2018multi}, which demonstrates that high performance for a simple baseline can be achieved in cases when passages have increasing lexical overlap with the question.

We note that this method potentially overestimates performance of a non-multi-hop model on HotpotQA, since there are some examples where many plausible answers are in the same sentence and require other context to resolve. However, these still form a minority in the dataset (see Table 3 of \newcite{yang2018hotpotqa}).



\begin{figure}[t]
    \centering
    \includegraphics[width=0.5\textwidth]{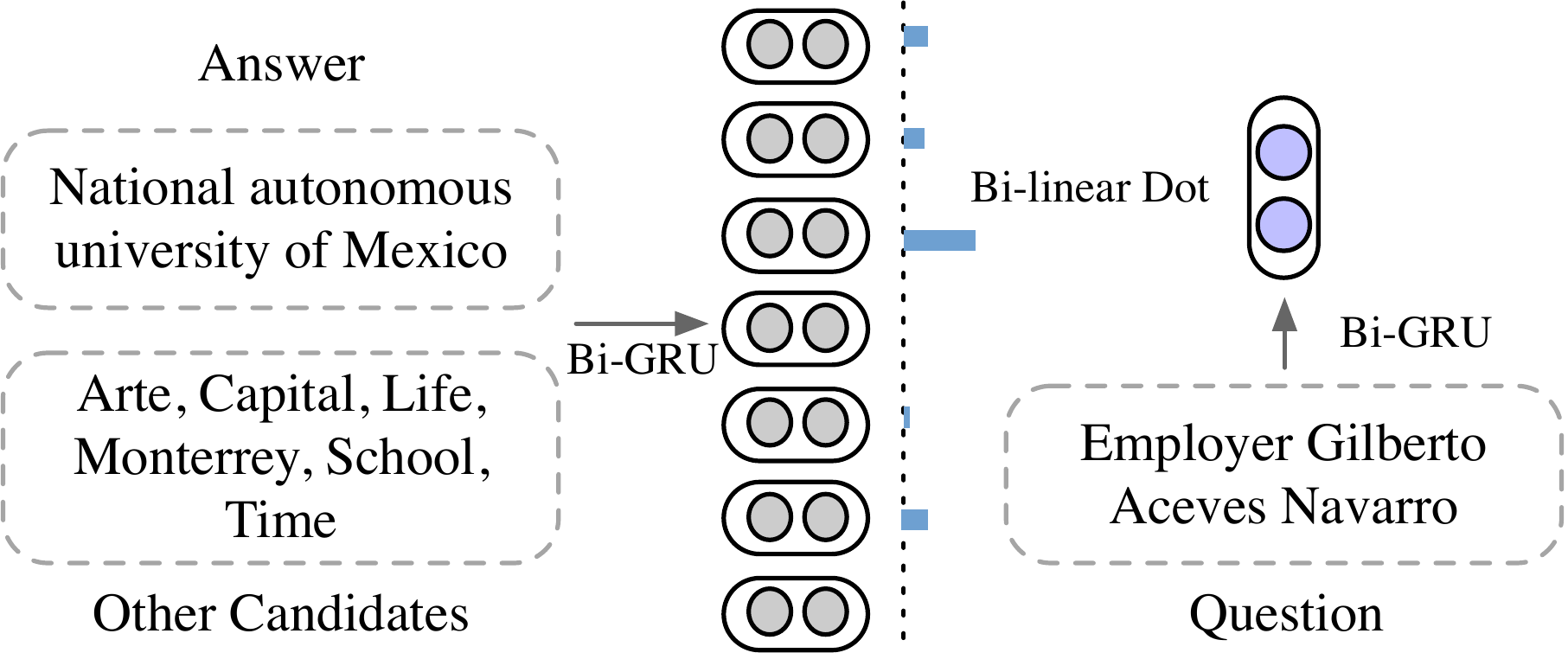}
    \caption{An example of question and candidates from WikiHop. Here we can see that among the candidates, only \emph{National autonomous university of Mexico} is an organization which could be Navarro's employer; the model may pick up on this entity typing.}
    \label{fig:dumb_baseline}
\end{figure}

\subsection{No Context Baseline}

The results of the previous section show that a model can identify correspondences between questions and answer sentences. One other pair of correlations we can study is suggested in the work of~\newcite{kaushik2018much}, namely examining question-answer correlations independent of the passage. We construct a ``no context'' baseline to verify whether it is possible to pick the correct answer without consulting the passage. In a similar fashion to the factored model, we encode the query $q$ and each answer candidate $c_i$ using a bi-GRU and once again compute a bilinear product between them to get the scores over candidates, making no reference to the document.

Results of this model on the multiple-choice WikiHop dataset are shown in Table~\ref{tab:noText_baseline}. Surprisingly, the no-context baseline achieves high performance, comparable to some recently-published systems, showing that WikiHop is actually possible to solve reasonably well without using the document at all. One possible reason for this is that this model can filter possible answers based on expected answer type~\cite{sugawara2018makes}, as shown in the example of Figure~\ref{fig:dumb_baseline}, or perhaps capture other correlations between training and test. This model substantially outperforms the unlearned baseline reported in the WikiHop paper \cite{welbl2018constructing} (38.8\%) as well as the BiDAF \cite{seo2016bidirectional} results reported there (42.9\%).





\begin{table}[t]
\centering
\small
\begin{tabular}{c | c |c | c}
\toprule

NoContext & Coref-GRU & MHQA-GRN & Entity-GCN  \\ \midrule
 59.70 & 56.00 & 62.80 & 64.80 \\
\bottomrule
\end{tabular}
\caption{The results of our no-context baseline compared with Coref-GRU~\cite{dhingra2018neural}, MHQA-GRN~\cite{song2018exploring}, and Entity-GCN~\cite{de2018question} on the WikiHop dev set. }
\vspace{-0.5cm}
\label{tab:noText_baseline}
\end{table}

\section{Span-based vs. Multiple-choice}
\label{section:experiment}

The no context model indicates that having multiple-choice questions may provide an avenue for a dataset to be gamed. In order to investigate the difference in multiple-choice vs.~span supervision while controlling for other aspects of dataset difficulty, we first recast each dataset in the other's framework, then investigate the performance of two models each of these settings.

To modify Hotpot to be multiple-choice, we randomly select 9 entities in all of the documents as distractors, and add the answer to make a 10-choice candidates set. To modify WikiHop to be span-based, we concatenate all documents and treat the first appearance of the answer mention as the gold span for training. Any answer occurrence is treated as correct for evaluation.


\begin{table}[t]
\small
\centering
\renewcommand{\tabcolsep}{1.3mm}
\begin{tabular}{ l | c  |  c   }
\toprule
 Dataset & HotpotQA-MC & WikiHop-MC \\ 
\midrule
Metric & Accuracy & Accuracy \\ \midrule
NoContext &  68.01 & 59.70\\
MC-BiDAF++ & 70.01 & 61.32\\
MC-MemNet & 68.75 & 61.80\\
\midrule
Span2MC-BiDAF++ & 76.01 & 59.85\\
\bottomrule
\end{tabular}
\caption{The performance of different models on the dev sets of WikiHop and HotpotQA. MC denotes using both the multiple-choice dataset and model. Span2MC means we train the model with span-based supervision and evaluate the model on a multiple choice setting. Our models only mildly outperform the no-context baseline in all settings.}
\vspace{-0.5cm}
\label{tab:mc_memnet_bidaf}
\end{table}

\subsection{Systems to Compare}




\paragraph{MemNet} Memory networks~\cite{weston2015towards} define a generic model class which can gather information from different parts of the passage. \citet{kumar2016ask} and \citet{miller2016key} have demonstrated its effectiveness in certain multi-hop settings. These models process a document over several timesteps. On the $i$th step, the model takes a question representation $q_i$, attends to the context representation $\mathbf{p}$, gets an attention distribution $\alpha_i$, computes a new memory cell value $m_i =\sum \alpha_i p_i$, then forms an updated $q_{i+1} = f(m_i,q_i)$. The final memory cell $m_T$ is used to compute a score $s_i = g(m_T,c_j)$ with the $j$th candidate representation $c_j$. We modify this architecture slightly using a standard hierarchical attention module~\cite{li2015hierarchical}.

We can also modify this architecture to predict an answer span -- we use the memory cell $m_T$ of the last step, and do a bi-linear product with the context representation $\mathbf{p}$ to compute a distribution over start points $P_{start} = \textrm{softmax}(\mathbf{p}W_{start}m_T)$ and end points distribution $P_{end} = \textrm{softmax}(\mathbf{p}W_{end}m_T)$ of the answer span, where $W_{start}$ and $W_{end}$ are two parameter matrix to be learned. We call this Span-MemNet.

\paragraph{BiDAF++} Recently proposed by~\newcite{clark2018simple}, this is a high-performing model on SQuAD. It combines the bi-directional attention flow~\cite{seo2016bidirectional} and self-attention mechanisms. We use the implementation described in~\newcite{yang2018hotpotqa}.

We can modify this model for the multiple-choice setting as well. Specifically, we use the start $P_{start}$ and end $P_{end}$ distribution to do a weighted sum over the context $\mathbf{p}$ to get a summarized representation $D_{start} = \sum P_{start_i} p_i$, $D_{end} = \sum P_{end_i} p_i$ of the context. Then we concatenate them to do a bilinear dot product with each candidate representation to get the answer score as we described for MemNet. We call this model MC-BiDAF++.




\begin{table}[t]
\small
\centering
\renewcommand{\tabcolsep}{1.3mm}
\begin{tabular}{ l | c c  | c c  }
\toprule
 Dataset & \multicolumn{2}{c|}{HotpotQA-Span} & \multicolumn{2}{c}{WikiHop-Span} \\ 
\midrule
Metric & EM & F1 & EM  & F1\\ \midrule
BiDAF++ (Yang+ 18) & 42.79 & 56.19 & $-$ & $-$ \\
\midrule
Span-BiDAF++ & 42.45 & 56.46  & 24.23 & 46.13\\
Span-MemNet & 18.75 & 26.11   & 13.54 & 19.23\\
\bottomrule
\end{tabular}
\caption{The performance of different models on the dev sets of WikiHop and HotpotQA. Span denotes using both span-based dataset and model. BiDAF++ denotes the performance reported in HotpotQA~\cite{yang2018hotpotqa}.}
\vspace{-0.5cm}
\label{tab:span-memnet_bidaf}
\end{table}
\subsection{Results}

Table~\ref{tab:mc_memnet_bidaf} and Table~\ref{tab:span-memnet_bidaf} show our results in the two settings. As a baseline on multiple-choice HotpotQA, we also test the no-context baseline, which achieves an accuracy of 68.01\%, around 10\% absolute higher than on WikiHop. Our candidates were randomly chosen, so this setting may not be quite as challenging as a well-constructed multiple-choice dataset. From Table~\ref{tab:mc_memnet_bidaf} and Table~\ref{tab:span-memnet_bidaf} we draw the following conclusions.

\paragraph{When trained and tested on multiple-choice datasets, our models do not learn multi-hop reasoning.} Comparing MC-BiDAF++ and MC-MemNet on the multiple-choice setting of both datasets as shown in Table~\ref{tab:mc_memnet_bidaf}, the models appear to have similar capabilities to learn multi-hop reasoning. However, looking at the no-context baseline for comparison, we find that it is only around 2\% lower than the two relatively more complex models. This indicates that much of the performance is achieved by ``cheating'' through the correlation between the candidates and question/context. Surprisingly, this is true even for HotpotQA, which seems stronger based on the analysis in Table~\ref{tab:decouple-test}.


\paragraph{Span-based data is less ``hackable'', but models still may not be doing multi-hop reasoning.} We then compare the results of Span-BiDAF++ and Span-MemNet on the span-based settings of both datasets, which are substantially different from the multiple-choice setting as shown in Table~\ref{tab:span-memnet_bidaf}. BiDAF++ substantially outperforms the MemNet on both datasets, indicating that BiDAF++ is a stronger model for multi-hop reasoning, despite being less explicitly designed for this task. However, this model still underperforms the Factored BiDAF model, indicating that it could just be doing strong single-sentence reasoning.

\begin{figure}[t]
    \centering
    \includegraphics[width=0.5\textwidth]{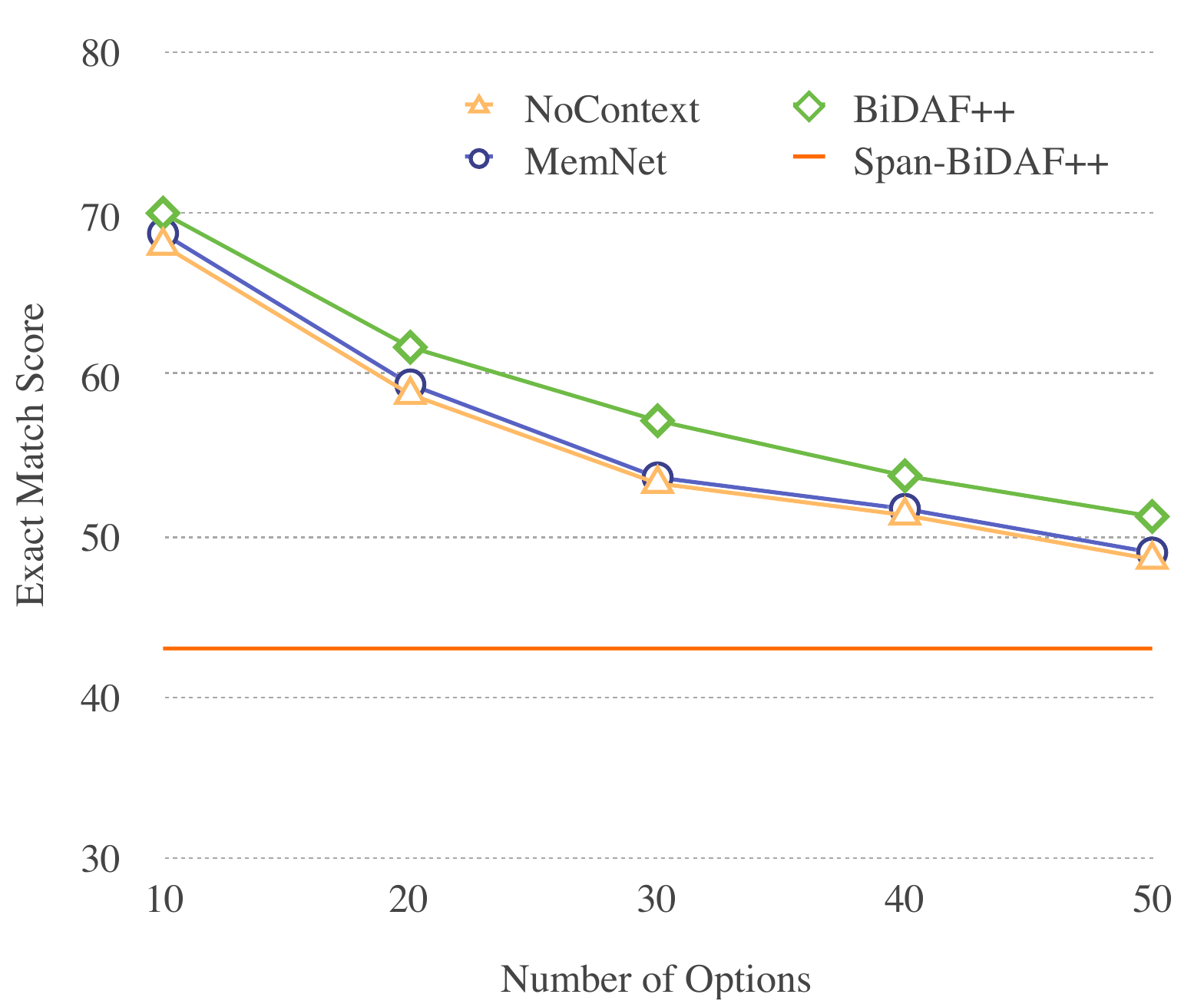}
    \caption{Performance of different options on HotpotQA-MC. Adding more options does not strengthen the model's ability of learning multi-hop reasoning. }
    \label{fig:various_options}
    \vspace{-0.5cm}
\end{figure}

\paragraph{Adding more options does not qualitatively change the multiple choice setting.} The span-based model requires dealing with a much larger output space than the multiple-choice setting. To test the effects of this, we conduct another experiment by making more spurious options on HotpotQA-MC using the method described in Section~\ref{section:experiment}. The results are shown in Figure~\ref{fig:various_options}. As we increase the number of options, we can see that the performance of all models drops. However, even with more options, the no-context baseline can still achieve comparable performance to the other two more complex models, which indicates that these models still aren't learning multi-hop reasoning in such a strengthened setting.

\paragraph{Span-based training data is more powerful.} To further understand the two different supervision signals, we conduct another experiment where we train using span-based supervision and evaluate on the multiple-choice setting. Specifically, during evaluation, we select all document spans that map onto some answer candidate, then max over the scores of all spans to pick the predicted answer candidate. The multiple choice options therefore filter the span model's predictions.

From the results in Table~\ref{tab:mc_memnet_bidaf}, we can see that Span2MC-BiDAF++ achieves higher performance compared to MC-BiDAF++ on HotpotQA and nearly comparable performance on WikiHop even with random span selection during training. This shows that with the span-based supervision, the model can learn at least the same thing as the multiple-choice and avoid ``cheating'' through learning question-candidate correspondences.

\section{Discussion and Conclusion}
There exist several other multi-hop reasoning datasets including WorldTree~\cite{jansen2018worldtree}, OpenBookQA~\cite{mihaylov2018can}, and MultiRC~\cite{khashabi2018looking}. These datasets are more complex to analyze since the answers may not appear directly in the passage and may simply be entailed by passage content. We leave a detailed investigation of these for future work.

For researchers working on the problem of multi-hop reasoning, we think the following points should be considered: (1) Prefer models using span-based supervision to avoid ``cheating'' by using the extra candidate information. (2) If using multiple-choice supervision, check the no-context baseline to see whether there are strong correlations between question and candidates. 
(3) When constructing a multi-hop oriented dataset, it would be best to do an adversarial test using a sentence-factored model to see whether multi-hop reasoning is really needed. Both HotpotQA and WikiHop contain good examples for evaluating multi-hop reasoning, but this evaluation is clouded by the presence of easily-solvable examples, which can confuse the learning process as well.


\section*{Acknowledgments}

This work was partially supported by NSF Grant IIS-1814522, NSF Grant SHF-1762299, a Bloomberg Data Science Grant, and an equipment grant from NVIDIA. The authors acknowledge the Texas Advanced Computing Center (TACC) at The University of Texas at Austin for providing HPC resources used to conduct this research. Results presented in this paper were obtained using the Chameleon testbed supported by the National Science Foundation. Thanks as well to the anonymous reviewers for their helpful comments.

\bibliography{naaclhlt2019}
\bibliographystyle{acl_natbib}

\end{document}